\documentclass{ifacconf}

\usepackage{graphicx}      
\usepackage{natbib}        

\usepackage{graphicx}
\usepackage{amsmath}
\usepackage{amsfonts}
\usepackage{algorithm}
\usepackage{algpseudocode}
\usepackage{cuted}
\usepackage{caption}
\usepackage{algorithm}
\usepackage{algpseudocode}
\begin{document}
\begin{frontmatter}

\title{Continual Test-Time learning \\ in Supply Chain} 


\author[First]{Guangya Hao}
\author[First]{Ge Zheng}
\author[Second]{Mike Shemko}
\author[Second]{Ed Prosser-Snelling}
\author[First,Third]{Alexandra Brintrup}

\address[First]{University of Cambridge, Cambridge, United Kingdom.}

\address[Second]{National Health Service (NHS), United Kingdom.}

\address[Third]{The Alan Turing Institute, London, United Kingdom.}
\address{E-mail: \{gh540, ab702\}@cam.ac.uk.}

\begin{abstract}                
Machine learning models in supply chain operations are often deployed under changing business conditions, where relationships learned from historical data may become unreliable and outcome labels are not immediately available. How deployed models can continuously adapt without external labels remains underexplored in supply chain AI research. This study investigates supply chain challenges that can be addressed through continual test-time learning and proposes a new adaptation method for evolving supply chain data environments. Evaluated on material delivery-delay prediction across four heterogeneous domains, the approach substantially improves performance after domain changes, demonstrating the potential of continual adaptation for maintaining model reliability under evolving supply chain conditions.
\end{abstract}


\end{frontmatter}

\section{Introduction}
Supply chain operations increasingly rely on predictions made under changing business conditions. Machine learning has been used to anticipate supplier disruptions and assess operational risks from historical records~\citep{brintrup2020disruptions,baryannis2019risks}. 
Reinforcement learning (RL) has also been applied to supply chain operations to support adaptive decision-making, particularly in inventory control and replenishment \citep{yan2022reinforcement,kosasih2022reinforcement}. However, changes in demand, supplier participation, transport services, and delivery commitments can alter the relationships captured in historical data~\citep{schoepf2025unlearning}. As these changes accumulate, deployed models may become less reliable, reducing their usefulness for inventory planning, transport allocation, and delivery risk management~\citep{gama2014drift}. A further challenge is that predictions are often required before their outcomes become available. In delivery-delay prediction, for example, an order must be assessed while fulfilment is still in progress, whereas its final delay status can only be determined later~\citep{saini2026leakage,alencar2026concept}. Supervised retraining therefore cannot immediately respond to change in operating conditions.

These challenges raise a practical question: \emph{how can a deployed model continuously adapt to changing operational conditions using incoming data when reliable outcome labels are not yet available?}

\begin{figure}[t]
\centering
\includegraphics[width=\columnwidth]{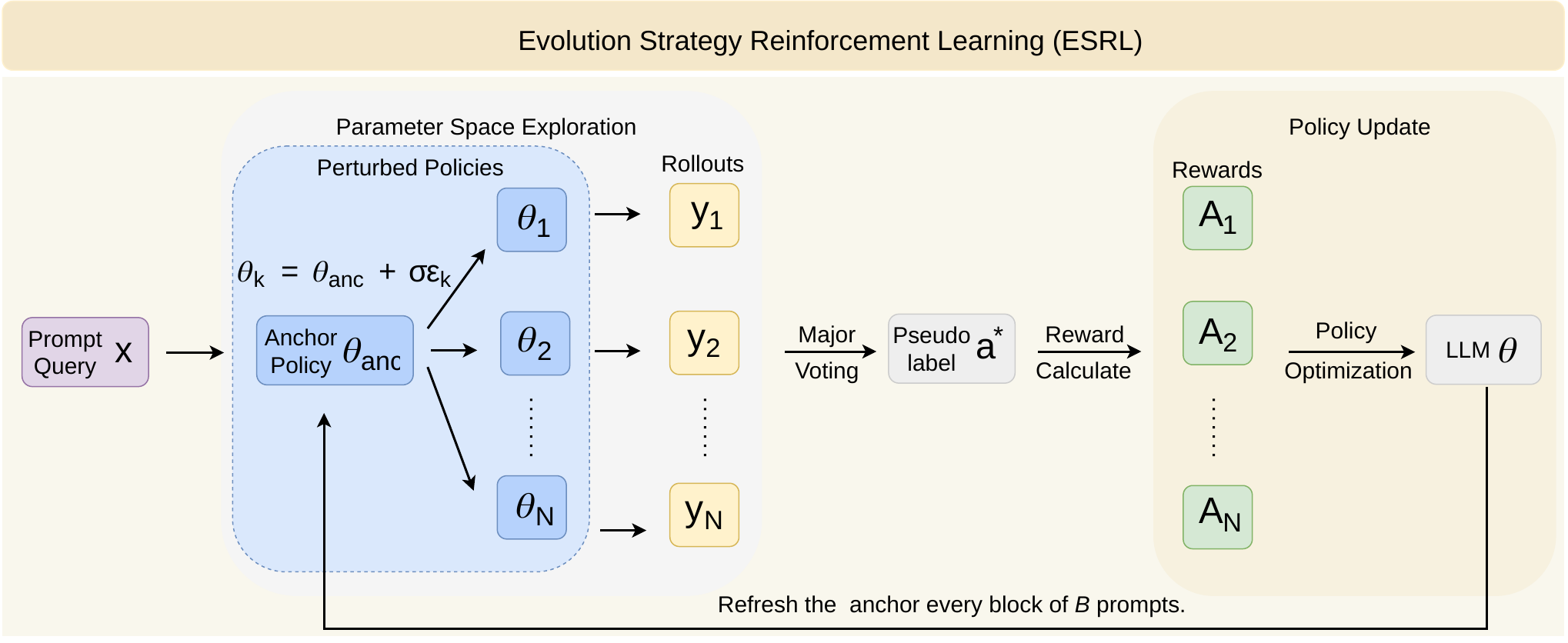}
\caption{Overview of ESRL. ESRL first constructs a population of perturbed policies around a shared anchor and generates one rollout from each policy. Majority voting provides label-free rewards, which are converted into group-relative advantages. The student is then updated with an advantage-weighted log-likelihood objective. The anchor remains fixed }
\label{fig:EvoStrategy}
\end{figure}

Continual test-time learning , which enables deployed models to continuously adapt to evolving data distributions using incoming unlabeled data, provides a framework for addressing this question by allowing models to update using data encountered during deployment~\citep{hao2026continual}. In this paper, we first examine supply chain operations (SCO) challenges that may benefit from continual test-time learning. We then introduce Evolution Strategy Reinforcement Learning (ESRL), which uses self-generated supervisory signals to support continual adaptation under changing operating conditions. Finally, we evaluate the proposed approach through a delivery-delay prediction case study across heterogeneous supply chain environments, examining how model performance changes under successive distribution shifts.

Our key contributions are:
\begin{itemize}
    \item We identify and discuss SCO challenges that can be addressed through continual test-time learning.
    \item We introduce ESRL as a continual adaptation method that leverages self-generated supervisory signals under changing operating conditions.
    \item We demonstrate the application of continual adaptation to delivery-delay prediction across heterogeneous supply chain environments.
\end{itemize}

\section{Continual Test-Time Learning and SCO Use Cases}

Continual test-time learning (CTTL) allows a deployed model to adapt using data encountered during inference, rather than remaining fixed after initial training~\citep{hao2026continual}. This is particularly relevant to supply chain operations (SCO), where changes in operational conditions and in supply chain relationships can make patterns learned from historical data less representative of current conditions~\citep{schoepf2025unlearning,senoner2023distributional,ezaki2021model}. At the same time, reliable outcome labels are often unavailable when predictions are required. For example, the final delivery status of an order can only be determined after fulfilment, although the prediction must be made earlier. The value of a new supplier contract must be predicted before budgets are planned. CTTL  provides a means of adapting deployed models using newly observed data without requiring immediate access to ground-truth labels. We discuss its potential application to SCO under two broad settings: changes in operational conditions and changes in supply chain relationships.

\subsection{Changes in Operational Conditions}

Supply chain operations are subject to changing demand, supply network configuration, order characteristics, fulfilment conditions, and disruption events~\citep{ezaki2021model,senoner2023distributional,brintrup2020disruptions}. As these conditions change, newly observed data can increasingly differ from the historical data used for model training, reducing the reliability of predictions under new operating conditions~\citep{senoner2023distributional,gama2014drift}. Three common forms of operational change are discussed below.

\textbf{Demand changes.}
Demand patterns change as market conditions and customer requirements evolve. Supply chain planning and forecasting models commonly rely on historical demand observations, but these models can become less representative when the underlying demand environment changes over time~\citep{ezaki2021model}.

\textbf{Order and fulfilment changes.}
Order characteristics and fulfilment conditions can also change over time. Differences between historical and forthcoming orders can introduce distributional shifts and reduce prediction performance when models are applied to new operational settings~\citep{senoner2023distributional}. This is particularly relevant to delivery-related tasks, where changes in order composition or fulfilment processes can change previously learned relationships.

\textbf{Disruptions and operational risks.}
Supply chains are also exposed to disruptions that change normal operating conditions. Machine learning has been used to predict supplier disruptions and delivery delays from historical supply chain data~\citep{brintrup2020disruptions,baryannis2019risks}. When the nature or frequency of disruptions changes, models trained on earlier observations can become less representative of current operations.

\subsection{Changes in Supply Chain Relationships}

Supply chains also change through the evolution of relationships among participating firms. Buyer--supplier relationships can develop, weaken, or be replaced as firms adjust sourcing strategies and inter-organisational arrangements~\citep{vanpoucke2014relationships,bygballe2017switching}. These changes can alter patterns captured in historical supply chain data and reduce the relevance of models trained on previous interactions.

\textbf{Supply chain configuration changes.}
The composition of a supply network changes as suppliers enter, leave, or are replaced. Supplier switching involves reducing or terminating an existing relationship while establishing an alternative one~\citep{bygballe2017switching}. Firms also continually evaluate and reselect suppliers, resulting in changes to the supplier portfolio over time~\citep{noorizadeh2021reallocation}. Historical observations associated with previous suppliers can therefore become less representative after the network changes.

\textbf{Supplier performance changes.}
The performance of an existing supplier can change even when the network structure remains unchanged. Variations in delivery reliability, capacity, disruption exposure, or product quality can alter the relationship between supplier characteristics and downstream outcomes~\citep{burkhart2024resilience}. Models based on historical supplier performance can consequently become less reliable when a supplier moves into a different operating regime.

\textbf{Buyer--supplier coordination changes.}
Relationships between existing partners can also evolve through changes in logistics responsiveness, information exchange, and the use of shared resources~\citep{vanpoucke2014relationships}. Such changes can affect how orders, commitments, and operational information translate into downstream outcomes, making relationships learned from earlier interactions less representative of current behaviour.

Taken together, changes in operational conditions and supply chain relationships can create a persistent mismatch between historical training data and the environment encountered after deployment. CTTL provides a mechanism for responding to this mismatch by updating models using newly observed test-time data~\citep{hao2026continual}. This is particularly relevant to SCO, where outcome labels for newly observed orders can become available only after a delay, limiting the ability of model to respond immediately to changing conditions~\citep{senoner2023distributional,saini2026leakage,alencar2026concept}.

\section{Continual Test-Time Learning with ESRL}

Test-time learning provides a way to adapt deployed models using data encountered after deployment when reliable outcome labels are unavailable. Recent approaches have increasingly relied on self-generated supervision to support such adaptation. Test-Time Reinforcement Learning (TTRL), for example, samples multiple responses from the current policy and uses majority voting to construct pseudo-labels and label-free rewards~\citep{zuo2025ttrl}. Subsequent methods improve the quality of these signals through more diverse exploration, reduced rollout collapse, or better regulation of the consensus used to construct pseudo-labels~\citep{zhou2025evolving,zhao2026echo,xu2026hittrl,hao2026memoey,wang2026ttpo,long2026self,hao2026self}.

While these methods mainly consider adaptation to a given test distribution, deployed models may encounter a sequence of changing distributions over time. Continual test-time learning extends test-time adaptation to this setting by allowing the model to update repeatedly as new test data arrive~\citep{hao2026continual}. This introduces an additional challenge: supervisory signals are repeatedly generated by policies that have themselves been modified by earlier updates. As the policy evolves, the source of its self-generated supervision evolves with it. Errors or increasingly homogeneous predictions can therefore be reinforced across successive distributions, amplifying pseudo-label errors, reducing exploration diversity, and contributing to catastrophic forgetting~\citep{hao2026continual}. Existing approaches may consequently improve performance on the current distribution while degrading capabilities acquired under previous distributions.

To address this problem, we employ Evolution Strategy Reinforcement Learning (ESRL), a continual test-time learning method that derives self-generated supervision from a population of nearby policies rather than repeatedly sampling trajectories from a single evolving policy~\citep{hao2026continual}. ESRL combines parameter-space population exploration with group-relative learning to obtain more diverse supervisory signals, while block-wise anchoring provides a stable reference during successive updates. These properties make ESRL suitable for SCO settings in which deployed models must continually adapt as operating conditions change over time.

\begin{figure}[t]
\centering
\includegraphics[width=\columnwidth]{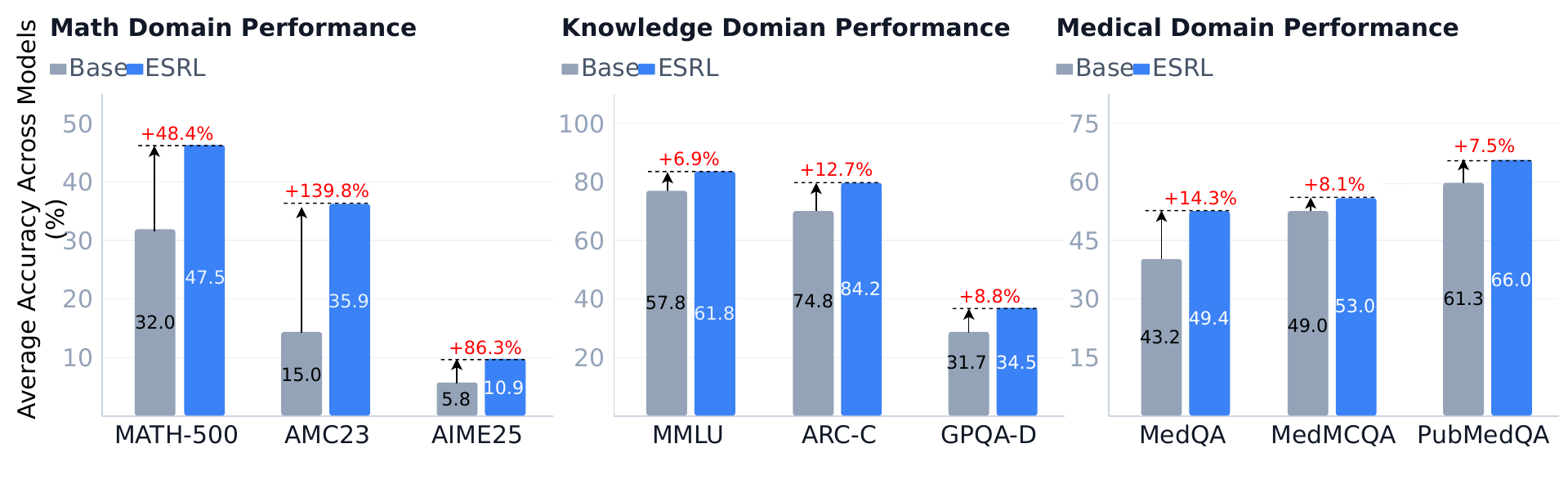}
\caption{Results reported in~\citep{hao2026continual} show that ESRL consistently improves performance across all evaluated domains compared with the base model.}
\label{fig:EvoStrategy}
\end{figure}

\subsection{Evolution Strategy Reinforcement Learning}

\begin{table*}[!t]
\centering
\caption{Performance of the HeteroAdapterNet source model
under direct transfer and domain-specific adaptation.
Changes ($\Delta$) are reported in percentage points.}
\label{tab:adaptation_results}
\setlength{\tabcolsep}{6pt}
\begin{tabular}{lccc|ccc|ccc}
\hline
& \multicolumn{3}{c|}{Accuracy (\%)}
& \multicolumn{3}{c|}{Balanced Accuracy (\%)}
& \multicolumn{3}{c}{AUROC (\%)} \\
\cline{2-10}
Domain
& Direct & Adapted & $\Delta$
& Direct & Adapted & $\Delta$
& Direct & Adapted & $\Delta$ \\
\hline
$D_2$
& 49.76 & 78.43 & +28.68
& 50.94 & 58.73 & +7.80
& 51.31 & 63.03 & +11.72 \\
$D_3$
& 67.65 & 91.99 & +24.34
& 54.40 & 82.69 & +28.29
& 66.18 & 90.36 & +24.19 \\
$D_4$
& 78.54 & 84.52 & +5.98
& 49.66 & 55.89 & +6.22
& 48.84 & 61.31 & +12.46 \\
\hline
\end{tabular}

\smallskip
\begin{minipage}{\textwidth}
\footnotesize
\end{minipage}
\end{table*}

\begin{figure}[t]
\centering
\includegraphics[width=\columnwidth]{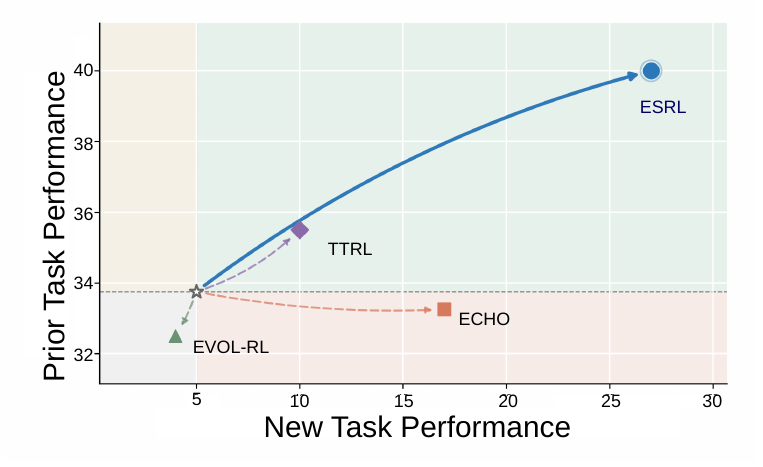}
\caption{Results reported in~\citep{hao2026continual} show that ESRL substantially reduces catastrophic forgetting and achieves the strongest final performance in continual learning compared with the evaluated baselines.}
\label{fig:EvoStrategy}
\end{figure}

ESRL maintains a central student policy with trainable parameters $\theta$ and an anchor policy with parameters $\theta_{\mathrm{anc}}$. Instead of repeatedly sampling multiple trajectories from the same policy, ESRL constructs a population of $N$ locally perturbed policies around the anchor. For each population member $k$, the perturbed parameters are given by

\begin{equation}
    \theta_k = \theta_{\mathrm{anc}} + \sigma \epsilon_k,
    \qquad
    \epsilon_k \sim \mathcal{N}(0,I),
\end{equation}

where $\sigma$ controls the perturbation magnitude. Given an unlabeled input $x$, each perturbed policy independently generates one rollout,

\begin{equation}
    y_k \sim \pi_{\theta_k}(\cdot \mid x),
    \qquad k=1,\ldots,N.
\end{equation}

This differs from conventional trajectory-space exploration, where multiple responses are sampled from a single policy. By perturbing the model in parameter space, ESRL obtains rollouts from multiple nearby policies, providing a broader source of self-generated supervision.

For each rollout $y_k$, an answer $a_k=\operatorname{Ans}(y_k)$ is extracted. The majority answer $a^{*}$ across the policy population is then used as the pseudo-label,

\begin{equation}
    a^{*}
    \in
    \arg\max_{a}
    \left|
    \left\{
    k:a_k=a
    \right\}
    \right|,
\end{equation}

and each rollout receives a label-free reward $ r_k$ according to its agreement with the population consensus,

\begin{equation}
    r_k=\mathbb{I}[a_k=a^{*}].
\end{equation}

The rewards are normalised within the population to obtain group-relative advantages $A_k$,

\begin{equation}
    A_k =
    \frac{r_k-\bar{r}}
    {\operatorname{Std}(r_1,\ldots,r_N)+\epsilon},
    \qquad
    \bar{r}=\frac{1}{N}\sum_{j=1}^{N}r_j,
\end{equation}

where $\epsilon$ is a small numerical constant. In this way, ESRL transforms agreement among independently perturbed policies into a self-generated learning signal without requiring external ground-truth labels.

The central student is subsequently updated using an advantage-weighted log-likelihood objective,

\begin{equation}
\begin{aligned}
\mathcal{L}_{\mathrm{ESRL}}(\theta)
=
\frac{1}{N}\sum_{k=1}^{N}
\Bigg[
&-A_k
\frac{1}{|y_k|}
\sum_{t=1}^{|y_k|}
\log \pi_{\theta}
(y_{k,t}\mid x,y_{k,<t}) \\
&+ \lambda \widehat{\mathrm{KL}}_k
\Bigg].
\end{aligned}
\end{equation}

where positive advantages increase the likelihood of consensus-supported trajectories, negative advantages suppress inconsistent trajectories, and the $\widehat{\mathrm{KL}}_k$ term constrains excessive deviation from the initial policy.
\begin{equation}
  \begin{aligned}
    \widehat{\mathrm{KL}}_k
    &=
    \frac{1}{|y_k|}
    \sum_{t=1}^{|y_k|}
    \left(
      e^{\delta_{k,t}}
      -
      \delta_{k,t}
      -
      1
    \right),
    \\
    \delta_{k,t}
    &=
    \log\pi_{\theta_0}
    \!\left(y_{k,t}\mid x,y_{k,<t}\right)
    -
    \log\pi_\theta
    \!\left(y_{k,t}\mid x,y_{k,<t}\right).
  \end{aligned}
  \label{eq:kl}
\end{equation}

Continual adaptation also requires the population to remain centred on a stable but sufficiently current policy. Updating the anchor after every input would cause the population to move continuously with the student, while keeping the same anchor throughout training would make it increasingly outdated. ESRL therefore processes inputs in blocks of $B$ examples. The anchor is fixed within each block and refreshed from the latest student at the next block boundary,

\begin{equation}
\theta_{\mathrm{anc}}^{(b+1)}
\leftarrow
\theta^{(b)}.
\end{equation}

This allows the population to explore around a stable reference within each block while progressively following the evolving student across successive environments. Together, population-based supervision and block-wise anchoring enable ESRL to adapt to newly encountered data while reducing error reinforcement and catastrophic forgetting. The detailed algorithm is provided in Appendix~\ref{algorithm}.

\section{Case Study: Continual Delivery-Delay Prediction}

For the SCO continual-adaptation case study, we construct a sequence of four heterogeneous supply chain domains using four publicly available datasets. The source domain $D_1$ is the DataCo Smart Supply Chain dataset~\citep{constante2019dataco}, representing order fulfilment operations. The subsequent target domains are the Olist Brazilian E-Commerce dataset ($D_2$)~\citep{olist2018ecommerce}, representing e-commerce parcel delivery; the SCOS Delivery History dataset ($D_3$)~\citep{usaid2018scms}, covering procurement and material shipments; and the Meituan-INFORMS-TSL Research Challenge dataset ($D_4$)~\citep{wang2025meituan}, representing large-scale on-demand delivery operations. After preprocessing, the four domains contain 59,490, 96,470, 3,393, and 568,545 valid order or shipment records, respectively, giving 727,898 observations in total.

The sequence $D_1\rightarrow D_2\rightarrow D_3\rightarrow D_4$ defines the experimental visitation order and is used to represent successive changes in supply chain conditions. The four datasets differ substantially in operational settings, feature distributions, business semantics, and delay prevalence, providing a challenging setting for evaluating continual adaptation across heterogeneous supply chain environments. We formulate a common material delivery-delay prediction task across all four domains. Before training, the datasets are converted into leakage-free order- or shipment-level representations and split temporally. Detailed preprocessing, feature harmonisation, and splitting procedures are provided in Appendix~\ref{app:data preprocessing}.

We adapt the network architecture of \citet{rebuffi2017learning} to handle heterogeneous supply chain features and refer to the modified model as HeteroAdapterNet. The model is first trained on $D_1$ to capture the underlying predictive relationships of the initial supply chain domain and is subsequently adapted to $D_2$, $D_3$, and $D_4$. Table~\ref{tab:adaptation_results} compares direct transfer of the original $D_1$ model with the final adapted model in each target domain.

 To assess the effectiveness of continual adaptation across
heterogeneous supply chain environments, we evaluate
predictive performance using three complementary metrics:
Accuracy, Balanced Accuracy, and Area Under the Receiver
Operating Characteristic Curve (AUROC).
Accuracy measures the overall proportion of correctly
classified instances:
\begin{equation}
\mathrm{Accuracy} =
\frac{TP+TN}{TP+TN+FP+FN}.
\end{equation}
where $TP$, $TN$, $FP$, and $FN$ denote true positives,
true negatives, false positives, and false negatives,
respectively. Since the proportion of delayed deliveries
varies across domains, we additionally employ Balanced
Accuracy to account for class imbalance by averaging
the recall of both classes:
\begin{equation}
\mathrm{Balanced\ Accuracy} =
\frac{1}{2}\left(
\frac{TP}{TP+FN}+
\frac{TN}{TN+FP}
\right).
\end{equation}
AUROC further evaluates the model's ability to
distinguish between delayed and non-delayed deliveries
across different classification thresholds:
\begin{equation}
\mathrm{AUROC} =
\int_0^1 \mathrm{TPR}
\left(\mathrm{FPR}^{-1}(u)\right)\,du.
\end{equation}
where $\mathrm{TPR}=TP/(TP+FN)$ and
$\mathrm{FPR}=FP/(FP+TN)$.
Higher values indicate better performance for all
three metrics.

The results show that $D_1$ serves as the source domain for initial model training, while $D_2$--$D_4$ evaluate adaptation to subsequently encountered supply chain environments. On $D_2$, adaptation substantially improves performance, increasing Accuracy from $49.76\%$ to $78.43\%$ and AUROC from $51.31\%$ to $63.03\%$. The largest improvement is observed on $D_3$, where Accuracy increases from $67.65\%$ to $91.99\%$, Balanced Accuracy from $54.40\%$ to $82.69\%$, and AUROC from $66.18\%$ to $90.36\%$. $D_4$ remains the most challenging target domain because of severe class imbalance and calibration shift. After adaptation and label-free rank calibration, Accuracy improves from $78.54\%$ to $84.52\%$ at the accuracy-oriented operating point, while the balanced operating point achieves $81.95\%$ Accuracy and $56.08\%$ Balanced Accuracy.

Overall, these results demonstrate that continual adaptation can substantially recover predictive performance across heterogeneous supply chain environments. They also highlight the importance of guarded model updates: ESRL can provide useful refinement when its self-generated supervision is reliable, but the update should be rejected when it degrades performance on the target-domain guard set.

\section{Conclusion}

This work explores the intersection of SCO and continual test-time learning, highlighting key challenges in SCO that can benefit from CTTL. We first identify supply chain settings in which evolving operational conditions and inter-organisational relationships can reduce the reliability of models trained on historical data. We then introduce ESRL as a continual adaptation mechanism and evaluate it through delivery-delay prediction across four heterogeneous supply chain domains. The findings demonstrate the potential of continual adaptation for maintaining performance of predictive models under changing supply chain conditions. Future research could extend continual test-time learning beyond delivery-delay prediction to other supply chain applications where operating conditions evolve and feedback is delayed or limited, such as demand forecasting, supplier risk assessment, and inventory or replenishment decision support. It would also be valuable to investigate how continual adaptation can be deployed in real-world supply chain systems where distribution shifts occur gradually and model updates must remain reliable over long operational periods.

\bibliography{ifacconf}             
                                                   







\appendix
\section{Dataset Preprocessing} 
\label{app:data preprocessing}

The four datasets are first converted into a common order- or shipment-level representation. DataCo records are aggregated by Order Id, Olist by \texttt{order\_id}, SCOS by ASN/DN shipment, and Meituan retains one final valid waybill for each order. Invalid or cancelled records are removed where applicable. Aggregation is performed before data splitting to prevent records belonging to the same order or shipment from appearing in different subsets. Identifiers, personally identifiable information, and variables only available after fulfilment, such as actual delivery status or completion time, are excluded from the model inputs.

To reduce differences in feature scales and availability across datasets, all domains are mapped to a common representation containing 66 numerical and six categorical features. Numerical features include within-domain standardised values, quantile-rank transformations, missing-value indicators, and temporal features. The categorical features represent provider, route, service, origin, destination, and product information and are mapped using fixed hash buckets. These fields provide a common representation across domains but do not imply exact semantic correspondence between datasets.

A common material delivery-delay label is constructed using the promised and actual delivery times. Let
\begin{equation}
L=t_{\mathrm{promised}}-t_{\mathrm{start}},
\qquad
\Delta=t_{\mathrm{actual}}-t_{\mathrm{promised}}.
\end{equation}

An order is classified as delayed when
\begin{equation}
y=
\mathbb{I}
\left[
\Delta>0
\;\land\;
\Delta\geq\max(\delta_d,0.10L)
\right],
\end{equation}
where $\delta_d$ accounts for the temporal resolution of each dataset. We use $\delta_d=24$ hours for DataCo, Olist, and SCOS, and one second for Meituan.

All datasets are split chronologically rather than randomly. For $D_2$--$D_4$, the earlier approximately $50\%$ of observations form the adaptation set and the later $50\%$ form the private test set. This preserves the temporal ordering within each domain and prevents future observations from being used during model adaptation.

\section{Algorithm}
\label{algorithm}

\begin{strip}
\centering

\setlength{\fboxrule}{0.6pt}
\setlength{\fboxsep}{6pt}

\fbox{%
\begin{minipage}{0.82\textwidth}

\captionof{algorithm}{Evolution Strategy Reinforcement Learning (ESRL)}
\label{alg:esrl}

\small
\begin{algorithmic}[1]

\Require Unlabeled prompts $\mathcal{D}$; student parameters $\theta$;
frozen reference policy $\pi_{\theta_0}$; population size $N$;
perturbation radius $\sigma$; block size $B$; agreement threshold $p_{\max}$;
KL weight $\lambda$; learning rate $\eta$

\For{each block $\mathcal{B}$ of at most $B$ prompts in $\mathcal{D}$}
    \State $\theta_{\mathrm{anc}} \gets \theta$
    \Comment{Synchronise the anchor with the student}
    \State Sample perturbation seeds $s_1,\dots,s_N$

    \For{$k=1$ to $N$}
        \State $\varepsilon_k \gets \varepsilon(s_k)$
        \State $\theta_k \gets \theta_{\mathrm{anc}}+\sigma\varepsilon_k$
        \State $\{y_k^{(x)}\}_{x\in\mathcal{B}}
        \gets \textsc{GenerateBatch}(\pi_{\theta_k},\mathcal{B})$
    \EndFor

    \For{each prompt $x\in\mathcal{B}$}
        \State $\{a_k\}_{k=1}^{N}
        \gets
        \{\mathrm{Ans}(y_k^{(x)})\}_{k=1}^{N}$

        \State $a^\star,\{r_k\}_{k=1}^{N}
        \gets
        \textsc{MajorityVote}(\{a_k\}_{k=1}^{N})$

        \If{\textsc{ValidityGate}$(\{a_k\}_{k=1}^{N},p_{\max})$ is false}
            \State \textbf{continue}
        \EndIf

        \State $\{A_k\}_{k=1}^{N}
        \gets
        \textsc{Standardise}(\{r_k\}_{k=1}^{N})$

        \State $\theta
        \gets
        \theta-\eta\nabla_{\theta}\mathcal{L}_{\mathrm{ESRL}}(\theta)$
       
    \EndFor
\EndFor

\end{algorithmic}

\end{minipage}%
}

\end{strip}

\end{document}